\documentclass{article}

\usepackage{arxiv}

\usepackage{cite}

\newcommand{\citep}[1]{\cite{#1}}
\newcommand{\citet}[1]{\cite{#1}}

\usepackage{amsmath,amssymb,amsthm}
\usepackage{dsfont,mathtools}
\usepackage{xspace}
\usepackage{epsfig}
\usepackage{algorithm}
\usepackage[noend]{algorithmic}
\usepackage{color}
\usepackage{url}
\usepackage{graphicx}
\usepackage{tikz}
\usepackage{float}
\usepackage{booktabs}
\usepackage{makecell}
\usepackage{multicol}
\usepackage{multirow}
\usepackage{csquotes}
\usepackage{wasysym}
\usepackage{longtable}
\usepackage{colortbl}
\usepackage{paralist}
\usepackage{hyperref}
\usepackage{ragged2e}

\definecolor{colorEAXR}{RGB}{27,147,108}
\definecolor{colorEAXRGPX}{RGB}{212,85,4}
\definecolor{colorLKHRIPT}{RGB}{105,100,170}

\title{One PLOT to Show Them All: Visualization of\\ Efficient Sets in Multi-Objective Landscapes}

\author{
  Lennart Sch{\"a}permeier\\
  Statistics and Optimization\\
  University of M{\"u}nster \\
  M{\"u}nster, Germany \\
  \texttt{schaepermeier@uni-muenster.de} \\
   \And
  Christian Grimme \\
  Statistics and Optimization\\
  University of M{\"u}nster \\
  M{\"u}nster, Germany \\
  \texttt{christian.grimme@uni-muenster.de}
  \And
  Pascal Kerschke \\
  Statistics and Optimization\\
  University of M{\"u}nster \\
  M{\"u}nster, Germany \\
  \texttt{kerschke@uni-muenster.de}
}

\begin{document}
\maketitle

\pagestyle{plain}
\thispagestyle{fancy}
\lfoot{\vspace*{-0.95cm}\rule{\columnwidth}{0.2pt}\\
\vspace*{-0.15cm}\footnotesize \begin{justify}This version has been accepted for publication at the \textit{16th International Conference on Parallel Problem Solving from Nature (PPSN XVI)}. Permission from the authors must be obtained for all other uses, in any current or future media, including re\-printing/re\-pub\-lishing this material for advertising or promotional purposes, creating new collective works, for resale or redistribution to servers or lists, or reuse of any copyrighted component of this work in other works.\end{justify}}\cfoot{}

\begin{abstract}

Visualization techniques for the decision space of continuous multi-objective optimization problems (MOPs) are rather scarce in research. For long, all techniques focused on global optimality and even for the few available landscape visualizations, e.g., \emph{cost landscapes}, globality is the main criterion. In contrast, the recently proposed \emph{gradient field heatmaps (GFHs)} emphasize the location and attraction basins of local efficient sets, but ignore the relation of sets in terms of solution quality.

In this paper, we propose a new and hybrid visualization technique, which combines the advantages of both approaches in order to represent local and global optimality together within a single visualization. Therefore, we build on the GFH approach but apply a new technique for approximating the location of locally efficient points and using the divergence of the multi-objective gradient vector field as a robust second-order condition. 
Then, the relative dominance relationship of the determined locally efficient points is used to visualize the complete landscape of the MOP. Augmented by information on the basins of attraction, this \emph{Plot of Landscapes with Optimal Trade-offs} (PLOT) becomes one of the most informative multi-objective landscape visualization techniques available.

\end{abstract}


\section{Introduction}
Traditionally, the visualization of optimization problems in decision space is one of the basic approaches to investigate challenges of so-called functional landscapes and to design basic algorithmic principles. Hence, low dimensional visualization is used in text books~\cite{Bey01,CC07} and algorithm research alike. For a single objective, each point in a continuous two-dimensional search space can be assigned with a function value which is interpreted as height. Overall, this leads to very natural notions of \emph{mountains} and \emph{valleys} for local maxima and minima, \emph{ridges} for discontinuities, as well as \emph{plateaus} for areas of equal height.

In evolutionary computation, many early theoretical results as well as later algorithmic concepts were first developed (and only successively generalized) by using low-dimensional visualizations of benchmark problems and their challenges. For continuous multi-objective optimization problems (MOPs), however, such straight-forward\rule{5pt}{0pt}visualization\newpage

techniques are not available. This is mainly rooted in the fact that MOPs comprise at least two contradicting objectives to be optimized simultaneously. Consequently, not a single or few global optimal solutions are sought but a set of optimal trade-off solutions -- the so-called \emph{Pareto set}. These solutions have as many objective function values (and thus height values) as objectives, which makes a standard landscape visualization infeasible. 

For few ($\le 3$) objectives, a classical visualization of the true or approximated set of efficient solutions, usually the Pareto front -- the Pareto set's image in objective space -- is used. However, by focusing purely on the objective space, one ignores all interaction effects from the MOP's input variables in the decision space.
Compared to the single-objective case, this is like reducing the entire landscape to the function values of its optimal solutions, and plotting them on a one-dimensional scale. 
Almost no information on the structural properties of the problem landscapes can be derived from this. Consequently, only little is known on MOP landscape properties and almost no algorithmic design is based on comparable insights like in the single-objective case. 

Although a straightforward visualization of MO landscapes is not available, there exist very few techniques for getting insights into these landscapes. The \emph{cost landscapes} proposed by Fonseca~\cite{Fonseca1995} use a dominance ranking approach to evaluate each point in decision space w.r.t.~the global optimal trade-offs. Although this delivers a kind of landscape in relation to the global optimum, it does not capture local optimal sets and their basins of attraction.
An alternative visualization approach that explicitly addresses locality has been proposed by Kerschke and Grimme~\cite{kerschke2017expedition,GrimmeKT2019Multimodality}. It produces (multi-objective) gradient field heatmaps (GFHs) using the Fritz-John (necessary) condition for identifying local optima~\cite{john2014extremum,miettinen2012nonlinear}. The GFHs show local basins and locally efficient sets but have two drawbacks: they do not provide a ranking of local sets w.r.t the global set and indicate local efficiency only indirectly by the height.
Within this work, we address the weaknesses of both approaches and contribute the following:
\begin{compactenum}
\item We propose a robust approach to determine locally efficient points explicitly. This includes a suitable second-order condition based on the divergence of the multi-objective gradient to confirm or exclude points that are considered to be locally optimal according to the first-order conditions.
\item Additionally, we combine and extend the two aforementioned state-of-the-art visualization approaches, i.e., the cost landscapes~\cite{Fonseca1995} and the gradient field heatmaps~\cite{kerschke2017expedition}. This leads to a far more informative visualization than any of these approaches taken by themselves offered before. We name this method \emph{Plot of Landscapes with Optimal Trade-offs (PLOT)}. Besides locality, global relations of local optima and respective basins can now be captured in a single PLOT. For two-dimensional problems, PLOT delivers a complete picture of the problem landscape that can be interpreted almost as seamless as a single-objective landscape, merely relating to the multi-objective gradient.
\end{compactenum}

The remainder of this work is organized as follows. Section~\ref{sec:background} summarizes the background by first providing fundamental notations and definitions that are needed later, and afterwards discussing the status quo on the visualization of MO landscapes. Section~\ref{sec:identification_local} describes the new methodology to determine locally efficient points, while Section~\ref{sec:visualization} describes the concept of merging the cost landscape approach and the gradient field heatmaps into PLOT. Finally, Section~\ref{sec:evaluation} evaluates our proposed PLOT approach by visualizing examples from current benchmark problems before Section~\ref{sec:conclusion} concludes the paper.

\section{Background}
\label{sec:background}

\subsection{Preliminaries on Multi-Objective Optimization}

For this work, we consider continuous MOPs $f: \mathbb{R}^p \rightarrow \mathbb{R}^k$ with search space parameter $p$, $k$ objectives and feasible search space $\mathcal{X} := [\mathbf{l}, \mathbf{u}] \subseteq \mathbb{R}^p$:
\begin{align}
    \begin{split}
        f(\mathbf{x}) = (f_1(\mathbf{x}), \dots, f_k(\mathbf{x})) \overset{!}{=} \min \qquad \text{ with } \qquad l_i \leq x_i \leq u_i, \; i=1, \dots, p.
    \end{split}
    \label{eq:mop}
\end{align}
The solution of a MOP is the set $\mathcal{X}^*\subseteq \mathcal{X}$ of Pareto-optimal trade-offs, i.e., all points $\mathbf{x}^*\in \mathcal{X}$ for which there exist no $\mathbf{x}'\in\mathcal{X}$ with $f_i(\mathbf{x}') \le f_i(\mathbf{x}^*)$ for all $i=1,\dots,k$ and $f_i(\mathbf{x}') < f_i(\mathbf{x}^*)$ for at least one $i$ (denoted as Pareto set). 
Thus, the aim of multi-objective (MO) optimization is to find all points in $\mathcal{X}$ that are not dominated by other points in the decision space. The image  $f(\mathcal{X}^*)$ is called the Pareto front.
\emph{Local efficiency of a point} is defined in analogy to locality in single-objective optimization: given a non-empty $\varepsilon$-neighborhood $B_{\varepsilon}(\mathbf{x})\subseteq\mathcal{X}$ around $\mathbf{x}$, no point $\mathbf{y}\in B_{\varepsilon}(\mathbf{x})$ dominates $\mathbf{x}$~\cite{Custodio2018}. Extending this definition to a set of locally efficient points, a \emph{local efficient set} is a set of points, which is not dominated by other points in their $\varepsilon$-neighborhood \cite{liefooghe:hal-01823666}. In a somewhat differentiated view, we can further discriminate different local sets, if we consider connected subsets of points as separate local efficient sets like it is done in~\cite{KerschkeWPGDTE16}. 

For the remainder of this paper, we will focus on two-dimensional bi-objective problems (i.e., $p=2$ and $k = 2$) to enable visual representations of the MOPs.
Although this may seem restrictive at first sight, it should be kept in mind that two-dimensional visualizations have substantially contributed to improving our understanding of the algorithmic search behavior in the single-objective case.
Also, we will adopt the notion of locally efficient sets from~\cite{KerschkeWPGDTE16} within this work.

The Fritz-John conditions are well known first-order conditions for continuous MOPs \cite{miettinen2012nonlinear}. Given a MO function $f$ defined as above, as well as inactive constraints, a first-order critical point $\mathbf{x}^*$ fulfills $\sum_{i=1}^{k} \lambda_i \nabla f_i(\mathbf{x}^*) = \mathbf{0}$ with $\lambda_i \geq 0$ for all $i=1,...,k$ and $\sum_{i=1}^k \lambda_i = 1$.
Based on this, a MO gradient $\nabla f (\mathbf{x})$ can be defined, which is zero if these conditions are satisfied, and which points towards a common descent direction of the objectives otherwise ($-\nabla f (\mathbf{x})$ for minimization). For two objectives, a definition for the MO gradient is given by the sum of the normalized single-objective gradients:
\begin{align}
    \nabla f(\mathbf{x}) = \nabla f_1(\mathbf{x}) / ||\nabla f_1(\mathbf{x})|| + \nabla f_2(\mathbf{x}) / ||\nabla f_2(\mathbf{x})||
    \label{eq:mog}
\end{align}

Following the MO gradient in a gradient-descent-like manner eventually leads into a local efficient set~\cite{GrimmeKEPDT2019SlidingToThe}, i.e., a (possibly connected) set of locally efficient points. Note that, 
if for a point $\mathbf{x}$ the length of one of its single-objective gradients is zero, i.e., $||\nabla f_1 (\mathbf{x})|| = 0$ or $||\nabla f_2 (\mathbf{x})|| = 0$, the point fulfills the necessary condition for a local optimum in the single-objective as well as in the MO case. We therefore define $\nabla f(\mathbf{x}) := 0$ for such points.

\subsection{Visualization of Continuous MOPs}
\label{sec:vis_background}

Benchmark problems are usually designed to test the capabilities and limitations of a broad spectrum of (optimization) algorithms~\cite{whitley1995building}. Aside from a pure performance comparison, the insights gained thereof are helpful for designing better algorithms. Here, `better' depends on various aspects such as the application or considered performance measure. Due to the different goals, a variety of test suites have been proposed over the years -- ranging from MOP~\cite{vanveldhuizen1999}, ZDT~\cite{Zitzler2000} and DTLZ~\cite{Deb2005} (which aim at posing challenges for MO evolutionary algorithms), over bi-objective BBOB \cite{tusar2016} (which extends the gold-standard test suite from single-objective optimization to the bi-objective case), up to more recent benchmark suites like the CEC 2019 test suite \cite{yue2019novel} (which emphasizes multimodality).

Although most test suites were designed with certain properties in mind, it remains questionable whether the contained MOPs actually meet them. So far, MOPs are predominantly visualized by means of their Pareto sets and/or fronts (see, e.g., \cite{Zitzler2000,yue2019novel}). Obviously, this is accompanied by an enormous loss of information, since all non-optimal points, and thus the information contained therein, are ignored.
The tools described in \cite{Tusar14phd,Tusar15tevc} offer a slight improvement over the very limited view at Pareto optimal points. However, the approaches described therein, such as the \emph{prosection} method, mainly focus on a dimensionality reduction of the objective space -- and thus do not permit drawing conclusions about the effects of the search space parameters on the objectives of the MOP.

To the best of our knowledge, there exist only two visualization techniques which provide a joint view at decision and objective space (and thus are helpful for engineering better algorithms and benchmark problems): the \emph{cost landscapes} by Fonseca~\cite{Fonseca1995} and the \emph{gradient field heatmaps (GFH)} by Kerschke and Grimme~\cite{kerschke2017expedition,kerschke2018search}. 
Both approaches have in common that they depict the decision space of two-dimensional MOPs and scalarize the information of the MOP's multiple objectives in a single height value (per point from the decision space). For both approaches, the decision space is discretized into a rectangular grid, and the grid's resolution naturally impacts the quality of the visualizations.

The height of a cost landscape is given by the so-called Pareto ranking, i.e., an integer value that gives the amount of points from the (discretized) decision space dominating the current point. Due to the usage of the (global) Pareto ranking, cost landscapes focus on \emph{global} optimality and thus are only able to reveal local structures, if the local fronts are close to parts of the Pareto front.

In contrast, the height of the GFHs is based on a MOP's gradient vector field. More precisely, for each point of the grid, the single-objective gradients (pointing to the closest optimum of the respective objective) are approximated using Eq.~\ref{eq:finite_diff}, and afterwards combined into a MO gradient as defined in Eq.~\ref{eq:mog}. Then, for each grid point, the gradient-based path towards the closest local efficient set is computed and the lengths of the MO gradients along the path are cumulated, defining the height of the GFH \cite{kerschke2017expedition}. Constructing the GFHs based on the paths towards the closest \emph{local} efficient set automatically provides insights into the local structure of the given MOP, as it depicts the attraction basins, as well as the local efficient sets contained therein. However, the focus on locality comes with the drawback that global relationships are hardly visible.

\begin{figure}[t!]
    \centering
    \includegraphics[width=0.32\textwidth, trim = 6mm 11mm 3mm 16mm, clip]{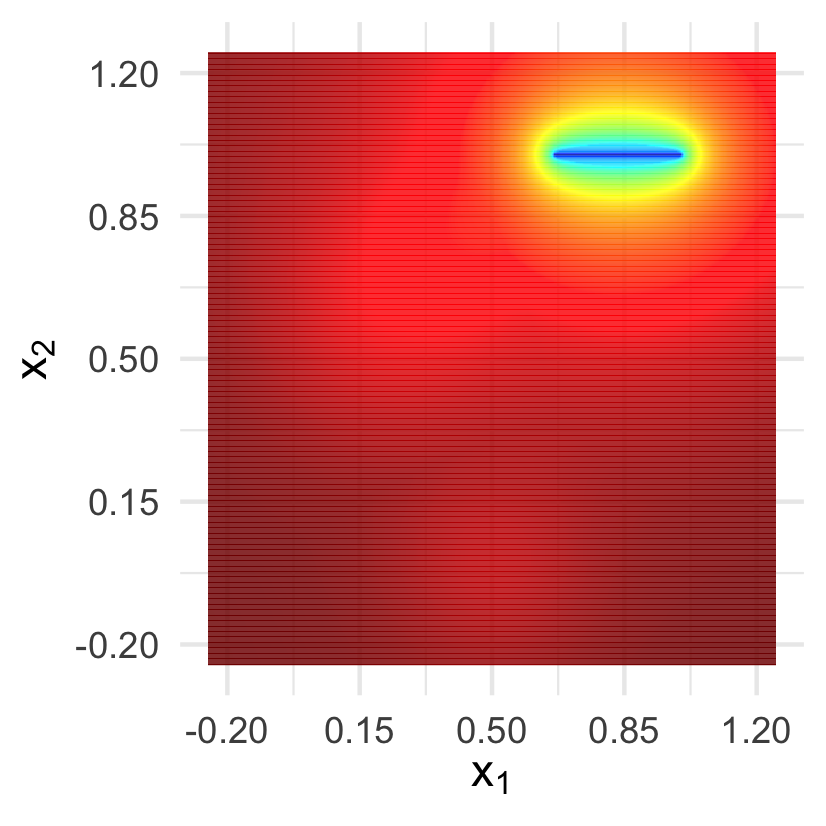}
    \hfill
    \includegraphics[width=0.32\textwidth, trim = 6mm 11mm 3mm 16mm, clip]{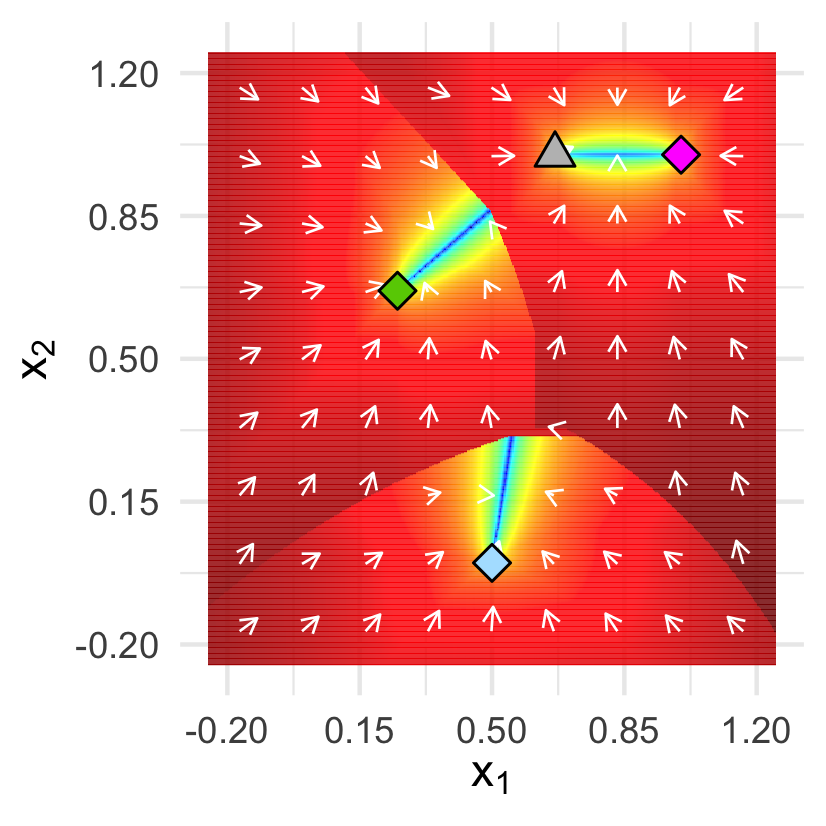}
    \hfill
    \includegraphics[width=0.32\textwidth, trim = 6mm 11mm 3mm 16mm, clip]{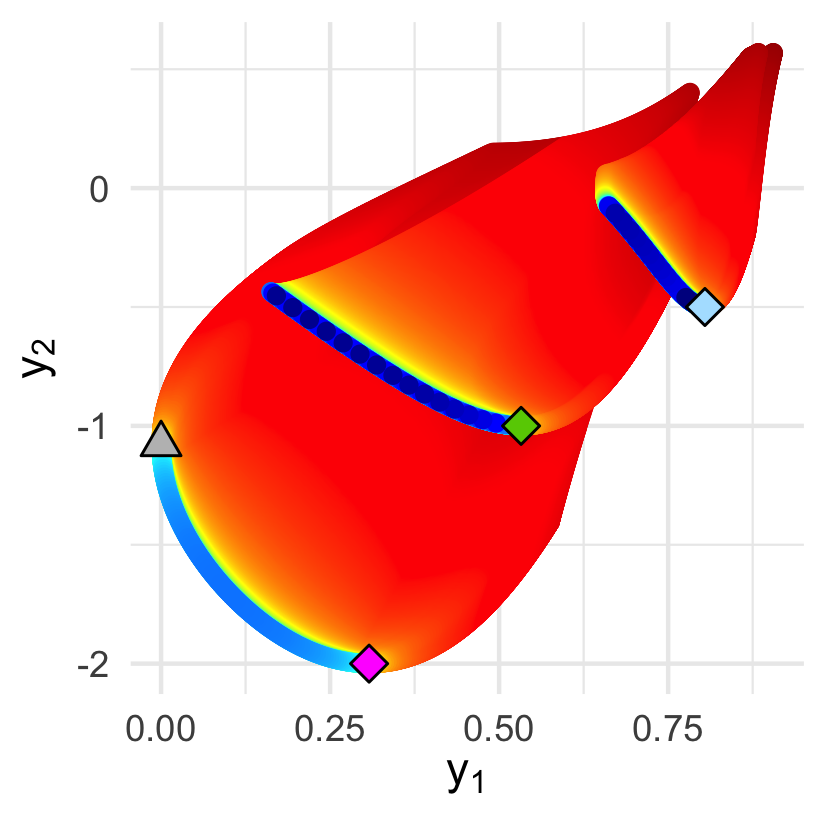}
    \caption{Exemplary comparison of the cost landscape (left) and the gradient field heatmap (middle) based on the bi-objective SGK function. The right image shows the objective space for the GFH and thus helps identifying the superposition and relationships of the three attraction basins (incl.~their associated efficient sets). The colors indicate the respective heights and change gradually from red (max.) to blue (min.).}
    \label{fig:viz-cost_vs_gfh}
\end{figure} 

Fig.~\ref{fig:viz-cost_vs_gfh} provides an exemplary visual comparison of the cost landscape and the GFH approach based on the bi-objective SGK function\footnote{$f(x_1, x_2)$ with $f_1(x_1, x_2) = 1 - \left(1 + 4 \cdot \bigl( (x_1 - 2/3)^2 + (x_2 - 1)^2\bigr) \right)^{-1}$ and\\ $f_2(x_1, x_2) = 1 - \max\{g_1, g_2, g_3\}$, whose subfunctions $g_1$, $g_2$ and $g_3$ are defined as $g_i(x_1, x_2, h, c_1, c_2) = h / \left(1 + 4 \cdot \bigl( (x_1 - c_1)^2 + (x_2 - c_2)^2\bigr) \right)$ with $h = 1.5$, $c_1 = 0.5$, $c_2 = 0$ (for $g_1$), $h = 2$, $c_{1} = 0.25$, $c_{2} = 2/3$ (for $g_2$), and $h = 3$, $c_{1} = 1 = c_{2}$ (for $g_3$)}, which combines a unimodal and a trimodal sphere function. The single-objective local optima are depicted by a grey triangle (for $f_1$), as well as green, blue, and pink diamonds (for $f_2$), respectively. The left image shows the corresponding cost landscape, in which the MOP's Pareto set is clearly visible -- in contrast to the local efficient sets, whose location can only be guessed by the shading of colors.
The GFH approach, given in the second image, shows the three attraction basins formed by the three competing optima of $f_2$, along with the corresponding vector field of MO gradients (white arrows). Moreover, all three (local) efficient sets are visible, and one can also identify two of them being non-global as their sets -- which start in the blue and green diamonds, resp. -- are abruptly cut by ridges between the current and the superseding attraction basins. While such a global ranking of the three efficient sets can be derived manually for this simple scenario, it is hard to realize for more complex MOPs (e.g., see bottom row of Fig.~\ref{fig:viz-comparison}).

\section{Identification of Locally Efficient Points}
\label{sec:identification_local}

Locally efficient points are an important part of MOP landscapes as they indicate where local Pareto fronts (or sets) are located, and thus, where local search strategies might get stuck \cite{Deb99}. However, state-of-the-art visualization approaches either do not feature them at all (e.g., the cost landscapes \cite{Fonseca1995}), or only show their locations indirectly (e.g., the gradient field heatmaps \cite{kerschke2017expedition}). Only when the location of the local efficient sets is known analytically for specific test problems -- like for DTLZ \cite{Deb2005} or MMF \cite{yue2019novel} -- they are represented in some visualizations. 

Here, we present an approach based on the estimated gradients of the MOP and the stability of the MO gradient vector field to locate all locally efficient points for a continuous MOP. We begin by detailing our computational approach for approximating the function and its derivatives, followed by a description of first- and second-order optimality conditions for locally efficient points and how they were implemented.

\subsection{Computational Approach}

As continuous functions can in principle be evaluated in infinitely many different points, an approximation of the function based on a finite set of evaluated points is required. For this purpose, we evaluate points on a rectangular grid of coordinates $(x_1^{j_1}, x_2^{j_2})$. Those coordinates are aligned equidistantly with a number of steps $n_1, n_2 \in \mathbb{N}$ and step sizes $s_i=(u_i - l_i) \cdot (n_i - 1)^{-1}$ per dimension, i.e.,
$
x_i^{j_i} = l_i + ({j_i} - 1) \cdot s_i, \text{ with } i = 1,2 \text{ and } j_i = 1,...,n_i
$
where $x_i^{j_i}$ denotes the coordinate of the ${j_i}$-th grid point in the $i$-th dimension.
Next, the rectangular grid of points is created by taking the cross-product of the one-dimensional coordinates $x_i^{j_i}$. The function $f$ is then evaluated for each of the points from the grid.

In each grid point, the respective derivative is approximated (per objective) using the finite differences method based on the neighboring coordinates of the point at hand. On the decision boundary, forward- and backward-differences are taken respectively, while the interior points are evaluated using central differences. With the function values of $f$ at grid point $(x_1^{j_1}, x_2^{j_2})$ denoted as $f(x_1^{j_1}, x_2^{j_2})$, the partial derivative of $f$ with regard to $x_1$ is then estimated by:
\begin{equation}
    \renewcommand{\arraystretch}{2}
    \frac \partial {\partial x_1} f(x_1^{j_1}, x_2^{j_2}) \approx
    \left\{
    \begin{array}{ll}
        \frac{1}{2 s_1} \cdot \left(f(x_1^{j_1 + 1}, x_2^{j_2}) - f(x_1^{j_1 - 1}, x_2^{j_2})\right), & \text{ for } 1 < j_1 < n_i \\
        \frac{1}{s_1} \cdot \left(f(x_1^{j_1 + 1}, x_2^{j_2}) - f(x_1^{j_1}, x_2^{j_2})\right), & \text{ for } j_1 = 1 \\
        \frac{1}{s_1} \cdot \left(f(x_1^{j_1}, x_2^{j_2}) - f(x_1^{j_1 - 1}, x_2^{j_2}) \right), & \text{ for } j_1 = n_1
    \end{array}
    \right.
    \label{eq:finite_diff}
\end{equation}
Derivatives for $x_2$ are calculated analogously. Compared to approximations with smaller step sizes, we did not observe noticeable changes in the visualization.

\subsection{First-Order Conditions}
\label{sec:first_order}

\begin{figure}[tb]
    \begin{center}
        \includegraphics[height=4cm]{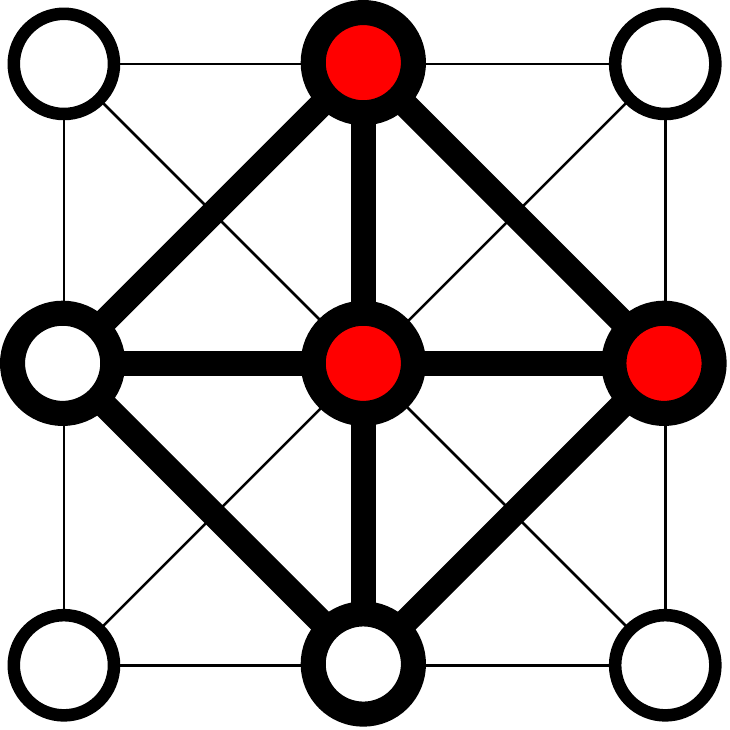}
        \hfill
        \includegraphics[height=4cm, trim = 0mm 1mm 3mm 0mm, clip]{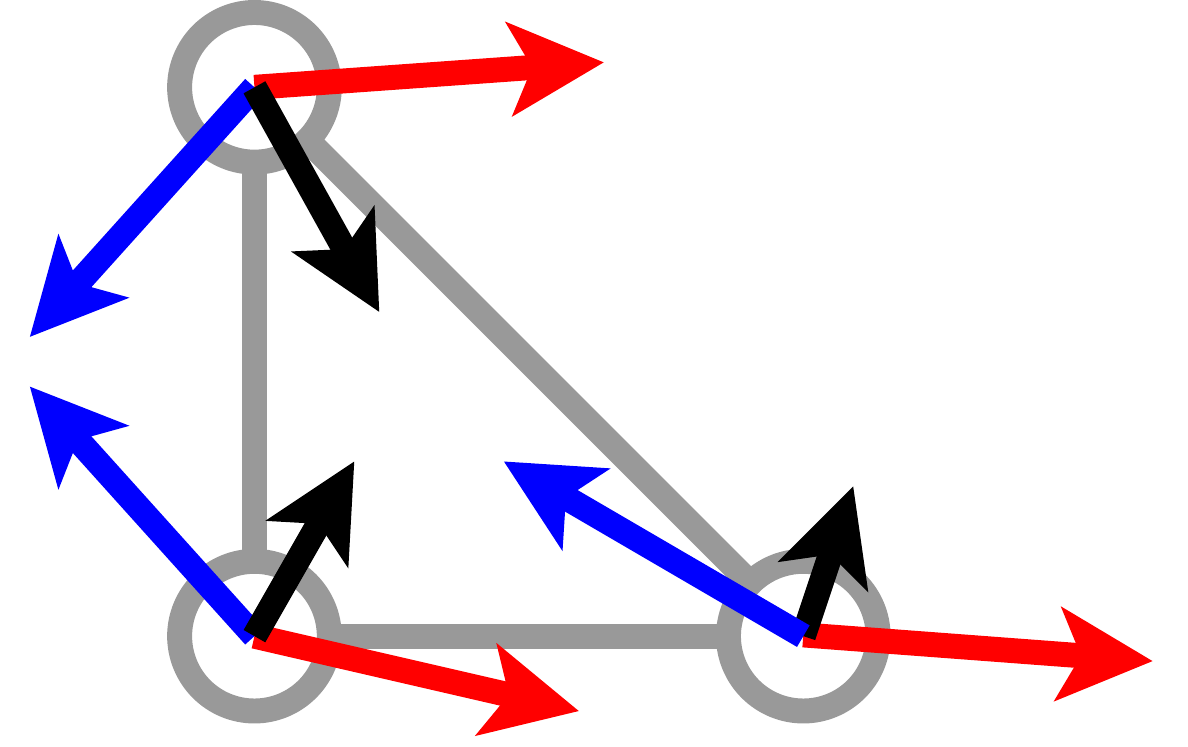}
        \includegraphics[height=4cm, trim = 2mm 1mm 0mm 0mm, clip]{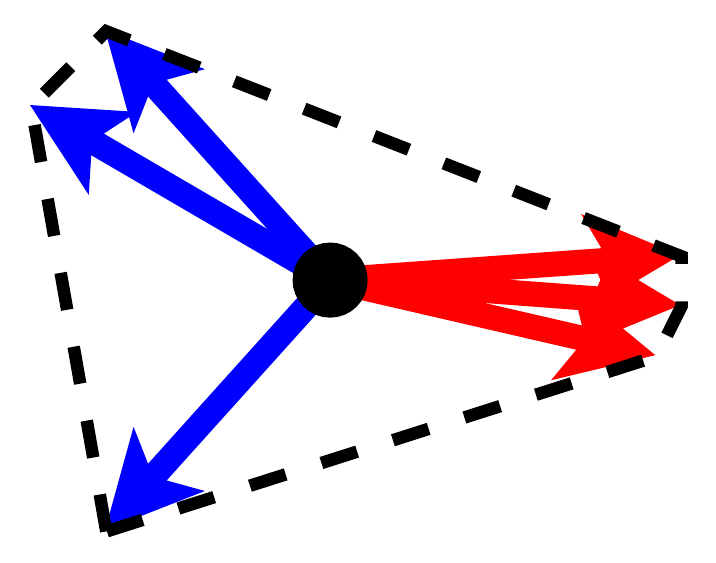}
    \end{center}
    \caption{Left: Example of four triangular neighborhoods considered while evaluating for critical points. The red nodes are an example for one particular neighborhood. Middle: For such a triangular neighborhood of points, considering the MO gradients (black), no critical points would be detected, as all of them point to the right as a common descent direction. However, it would be reasonable to expect that due to the changes in direction of the single-objective gradients (red and blue respectively, pointing towards descent directions), there would be a local efficient set in between. Right: Our approach jointly considering the combination of all single-objective gradients from all corner points correctly reports the neighborhood to contain a locally efficient point.} \label{fig:gradients}
\end{figure}

Although the MO gradient is capable of capturing the local efficiency of a point in the decision space, it is not sufficient on its own to capture all critical points while approximating them in the grid-based decision space. 
In particular, if one of the single-objective gradients changes much faster than the other in the neighborhood of a local efficient set, the MO gradient alone may misleadingly fail to recognize some of the critical points (as shown schematically in Fig.~\ref{fig:gradients}). 

Thus, we present an approach that improves the status quo in detecting all critical points within the MOP's grid. We consider triangular neighborhoods of grid points and jointly consider the gradients of all constituent single-objective functions. If the convex hull of the gradients encloses the origin (see right image of Fig.~\ref{fig:gradients}), we presume a critical point \emph{somewhere} in the interior of the triangle.

An illustration of the neighborhood, which is used for detecting the critical points, is given in the left image of Fig.~\ref{fig:gradients}. For each grid point, four triangular neighborhoods are evaluated. Each of those triangles consists of the point itself, as well as one horizontal and one vertical neighbor of that point. If a critical point is detected for a triangle, all of its corner points are considered being critical.

In addition, points along the decision boundary require special attention. So far, in the gradient field heatmaps, all (boundary) points for which the MO gradient points ``out'' of the feasible decision space were considered locally efficient. However, even in that case, it might still be possible to follow a descent direction for all objectives when moving along the decision boundary. 
We resolve this issue by considering boundary points, their adjacent boundary points and their common descent directions along the decision boundary all together. 
If no common descent direction for all objectives is found, the respective pair of boundary points is considered critical. Further, we rotate the MO gradient at these points to point into the common descent direction along the boundary, which is helpful for further visualizations with the gradient field heatmap (see Sec.~\ref{sec:visualization}).

\begin{figure}[tb]
    \centering
    \phantom{a}
    \hfill
    \includegraphics[width=0.4\textwidth, trim = 0mm 10mm 0mm 15mm, clip]{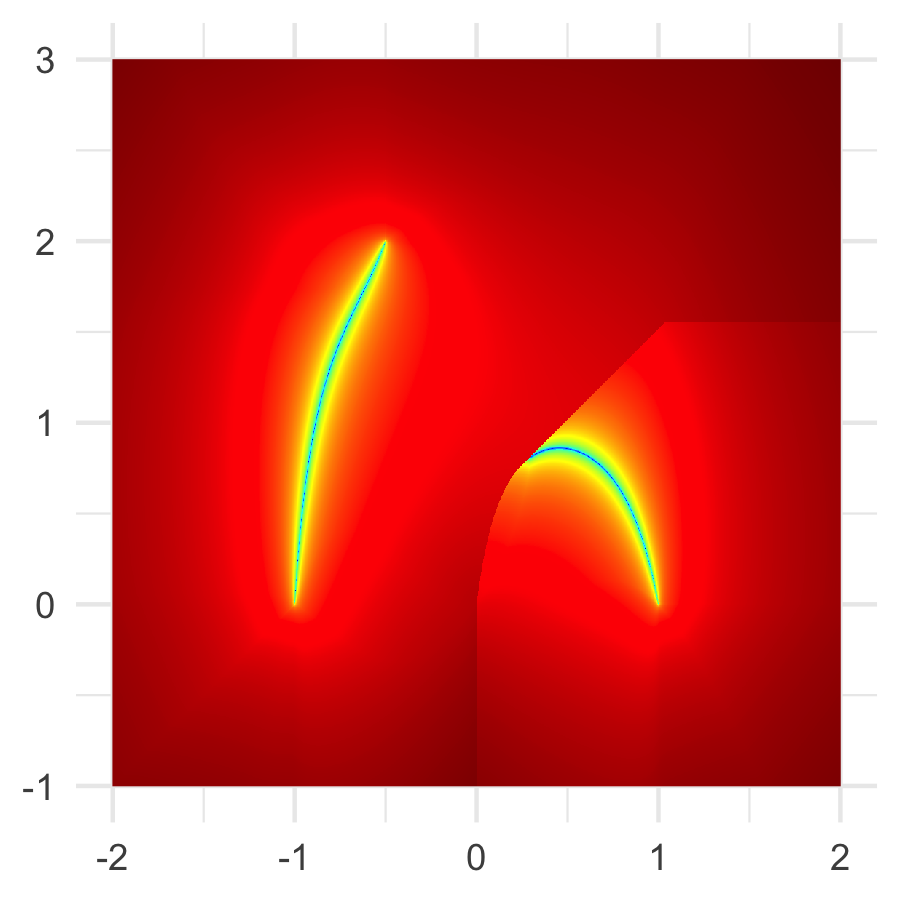}
    \hfill
    \includegraphics[width=0.4\textwidth, trim = 0mm 10mm 0mm 15mm, clip]{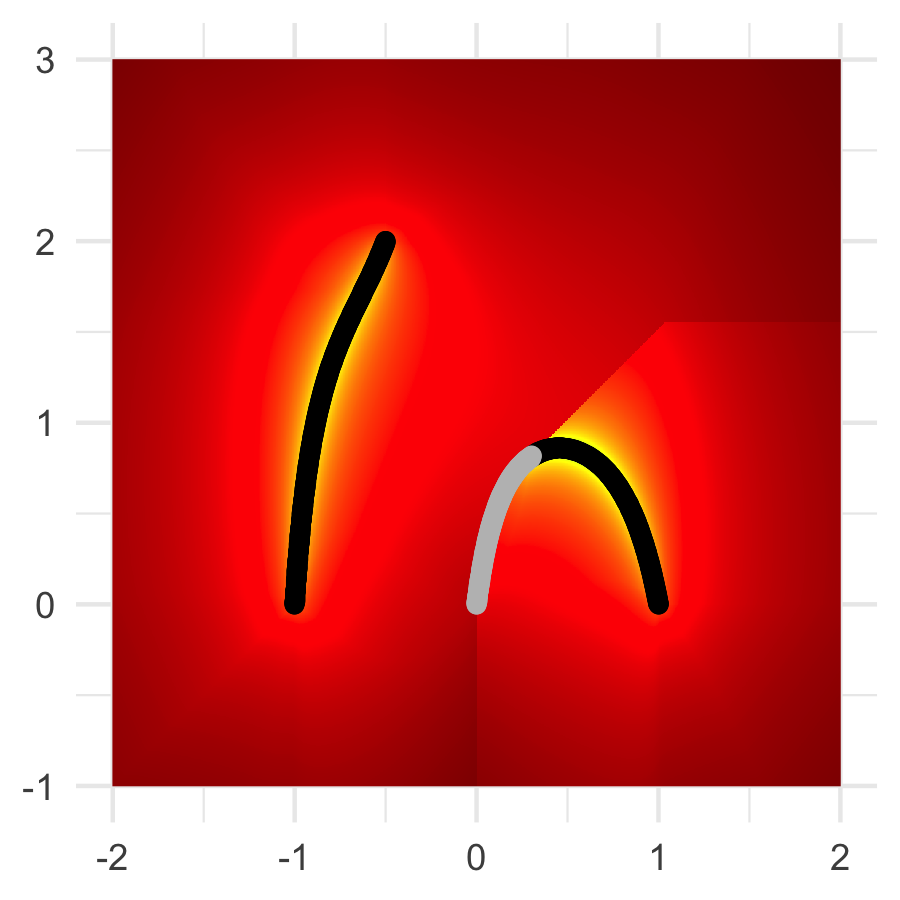}
    \hfill
    \phantom{a}
    \caption{GFH visualization and location of all interior critical points for the Aspar function $f(x_1, x_2) = (x_1^4 - 2 x_1^2 + 2 x_2^2 + 1, (x_1+0.5)^2+(x_2-2)^2)$. Even in this very simple problem, some critical points (gray) do not belong to the efficient sets (black) but are part of the landscape's ridges, emphasizing the need for a second-order criterion.} \label{fig:aspar-gfh}
\end{figure}

When only aiming at the identification of locally efficient points in the decision space, one needs to be aware that not all critical points are necessarily locally efficient. An example of this is given in Fig.~\ref{fig:aspar-gfh}, which shows the GFH of a simple MOP and all of its critical points. Note that some of the critical points do not belong to a local efficient set, but are rather part of the ridges between two adjacent basins of attractions. To reliably extract the locally efficient points from the set of critical points, a second-order condition is required.

\subsection{Second-Order Conditions}
\label{sec:second_order}

Analogously to single-objective functions satisfying first-order optimality conditions, critical points in the multi-objective sense cannot just be local minima, i.e., locally efficient points, but also local maxima, as well as saddle points.

This highlights the need for the consideration of a second-order condition to distinguish identified critical points into locally efficient points and others. There are second-order conditions for the continuous case \cite{miettinen2012nonlinear}, however, using the grid approximation required for our approach, these proved to be too unstable for efficient use to discern between the different types of critical points.
Motivated by the gradient field heatmaps, which indicate that the MO gradient captures the local search behaviour well, we derive our second-order condition based on properties of the MO gradient vector field.

The MO gradient defines a vector field over the decision space that can be analyzed w.r.t.~its stability at the critical points. A point is considered asymptotically stable, if after a small perturbation, following the vector field to the closest critical point, one stays within a small neighborhood of the original point. This property is well studied in the field of autonomous differential equations and can be analyzed by a linear approximation of the vector field at the critical point using its Jacobian \cite{blanchard2012differential}.\footnote{This presumes that the MO gradient field can be approximated by a linear function in the considered point. For differentiable MOPs this is a reasonable assumption, but we observed that it works well for our approach in general.} If the real part of all (potentially complex-valued) eigenvalues of the Jacobian is negative, the point is considered stable.

In the MO gradient field, however, we generally deal with degenerated critical points, for which (at least) one eigenvalue is zero, associated with the eigenvectors pointing along the local efficient set. This can pose problems with the numeric approximation that we require for our approach. Luckily, for the 2D case it is sufficient to consider the trace of the Jacobian, also known as the divergence in the context of vector fields, to determine asymptotic stability. Intuitively, the divergence is a measure for the ``ingoingness'' or ``outgoingness'' of the vector field at a given point, and it is numerically more stable and efficient to calculate than computing all eigenvalues of the Jacobian. Thus, the divergence of a 2D vector field $\mathbf{V}: \mathbb{R}^2 \rightarrow \mathbb{R}^2$ with $\mathbf{V} (\mathbf x) = (V_{x_1} (\mathbf x), V_{x_2} (\mathbf x))$ is given by:
\begin{align}
    \mathbf{div} (\mathbf V (\mathbf x)) = \frac \partial {\partial {x_1}} V_{x_1}(\mathbf x) + \frac \partial {\partial {x_2}} V_{x_2}(\mathbf x).
\end{align}
In summary, for minimization as defined in Eq.~\ref{eq:mop}, if an interior critical point $\mathbf{x}$ fulfills $\mathbf{div} (-\nabla f(\mathbf{x})) < 0$, it is a stable critical point in the MO gradient field, and thus locally efficient. Note that in the degenerated case, where all points in a given neighborhood are locally efficient, the divergence is zero.

Thus, to assess whether a critical point is locally efficient, we consider the divergence of each set of points that were jointly considered critical w.r.t.~the first-order conditions (see Sec.~\ref{sec:first_order}). Only if all three evaluated points have non-positive divergence, we regard them as locally efficient. The divergence for each grid point is estimated using the grid-based finite differences method (see Eq.~\ref{eq:finite_diff}).

Again, the critical points along the decision boundary require special treatment. Here, we do not use the divergence to distinguish different types of critical points. Instead, only if the MO gradients of the considered boundary points are pointing ``outwards'' or along the decision boundary, a set of critical points is considered locally efficient.

\section{Visualizing Local and Global Structures of MOPs}
\label{sec:visualization}

The previous section introduced a novel and reliable approach to approximate the location of all locally efficient points within the MOP's continuous decision space (based on a rectangular grid of evaluated points). 
This explicit knowledge of the location of the locally efficient points not only allows us to show them in the decision space, but also to extract information about their relative dominance relationship. This means, we utilize Pareto ranking, which also serves as the basis for the cost landscapes \cite{Fonseca1995}, but limit ourselves to the locally efficient points -- resulting in an enormous speed-up compared to a ranking of \emph{all} grid points.

Ultimately, this leads to a unique visualization that not only shows the locations of locally efficient solutions, but also provides information about their global optimality. In addition, we enhance our visualization with a gray-scaled version of the corresponding GFH in the background, which preserves additional information about the basins of attraction (e.g., their shapes and sizes). Along the boundary points, we modify the MO gradient as described in Sec.~\ref{sec:first_order}. Also, all locally efficient points determined by our more stable detection method (see Sec.~\ref{sec:identification_local}) can automatically be reused ``for free'' within the generation of the GFHs. Both modifications further improved the visualization quality of the GFHs. 
\begin{center}
\centering
    \begin{tabular}{|p{16.0cm}|}
    \hline
    \rule{0pt}{11pt}Ultimately, this results in a single \emph{\underline{P}lot of the \underline{L}andscape with \underline{O}ptimal \underline{T}rade-offs} (PLOT), which jointly illustrates three types of landscape characteristics: (1) \emph{local efficient sets}, (2) the \emph{global optimality} of their respective solutions, and (3) the \emph{basins of attraction} associated with the respective efficient sets.\\[-0.95em]
    \rule{0pt}{1pt}\\
    \hline
    \end{tabular}
\end{center}
\begin{figure}[t!]
    \centering
    \includegraphics[width=0.32\textwidth, trim = 0mm 2mm 2mm 4mm, clip]{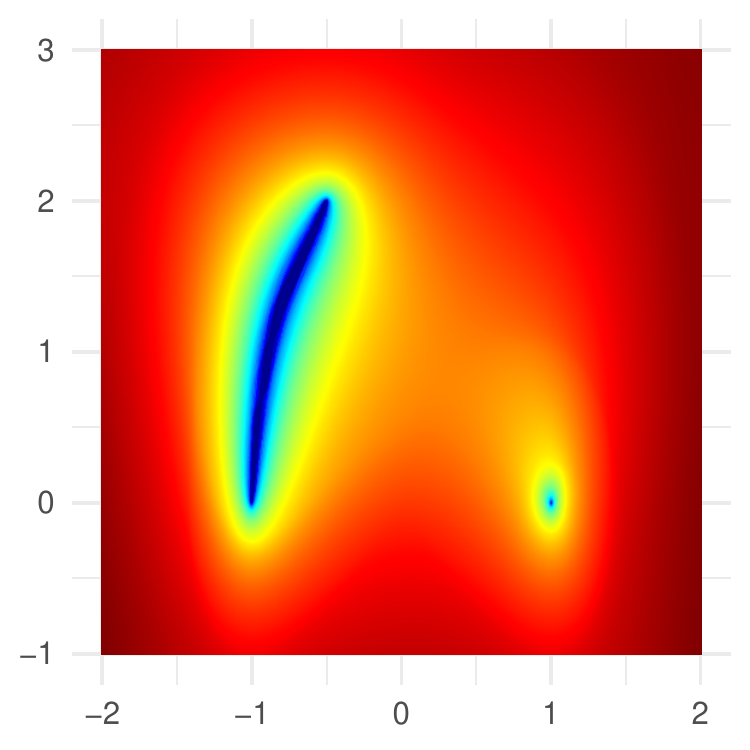}
    \includegraphics[width=0.32\textwidth, trim = 0mm 2mm 2mm 4mm, clip]{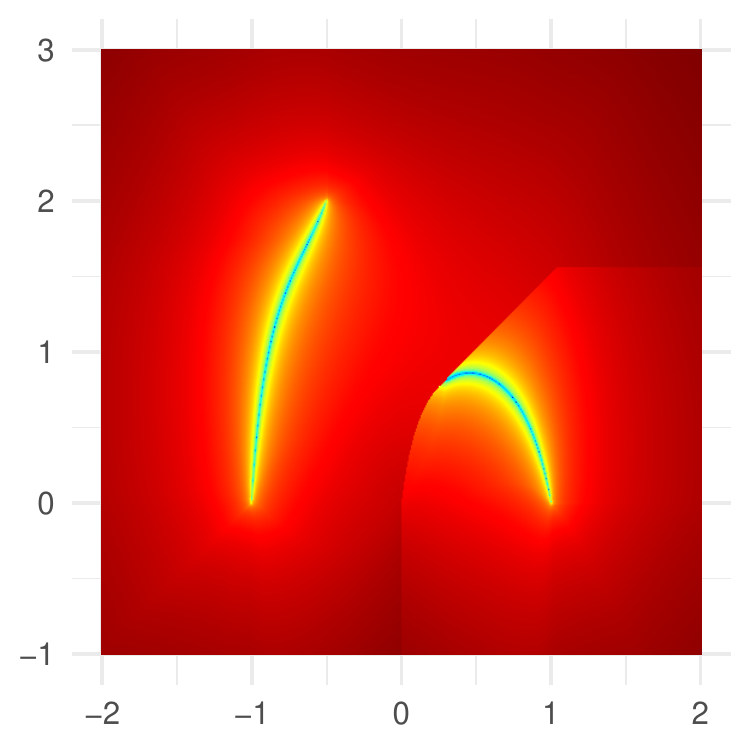}
    \includegraphics[width=0.32\textwidth, trim = 0mm 2mm 1mm 4mm, clip]{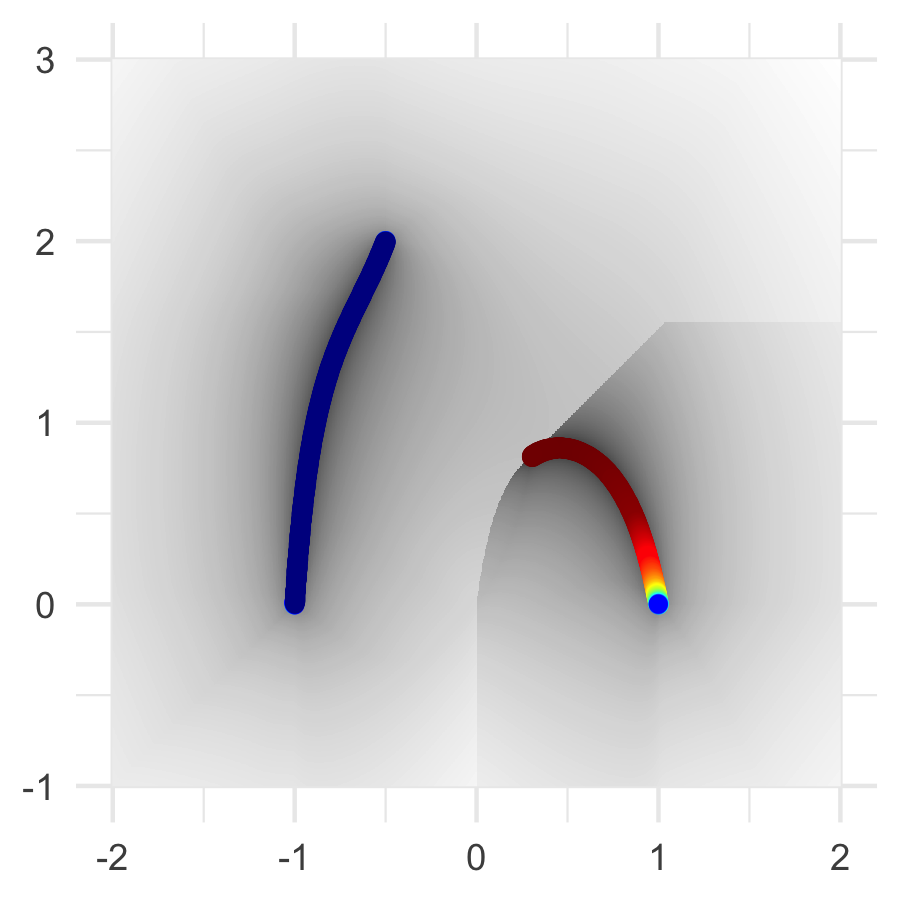}
    \includegraphics[width=0.32\textwidth, trim = 0mm 2mm 2mm 4mm, clip]{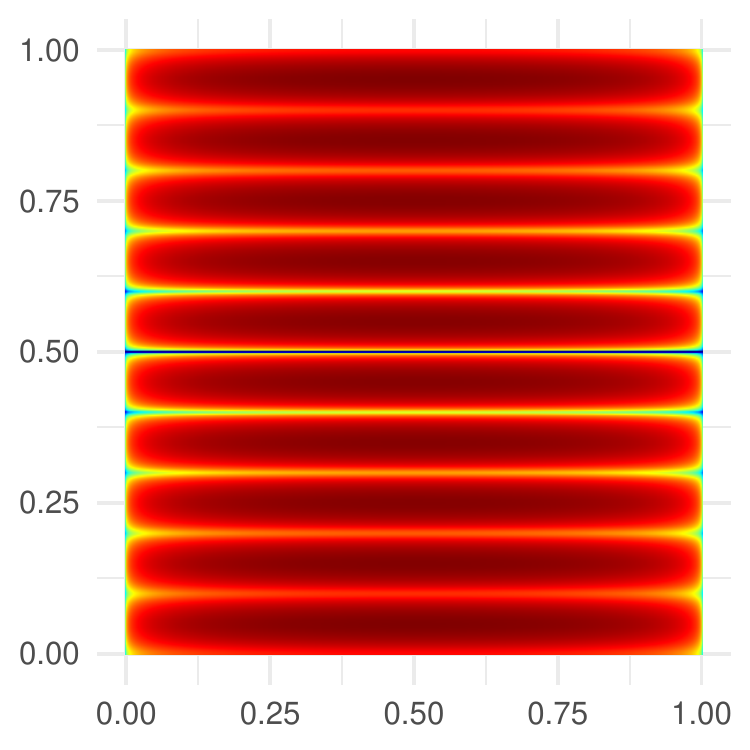}
    \includegraphics[width=0.32\textwidth, trim = 0mm 2mm 2mm 4mm, clip]{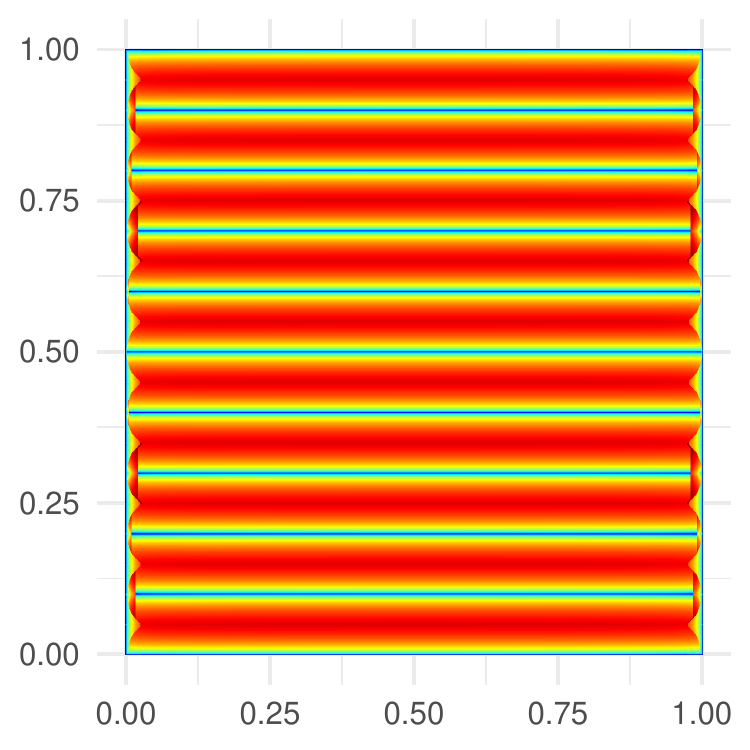}
    \includegraphics[width=0.32\textwidth, trim = 0mm 2mm 1mm 4mm, clip]{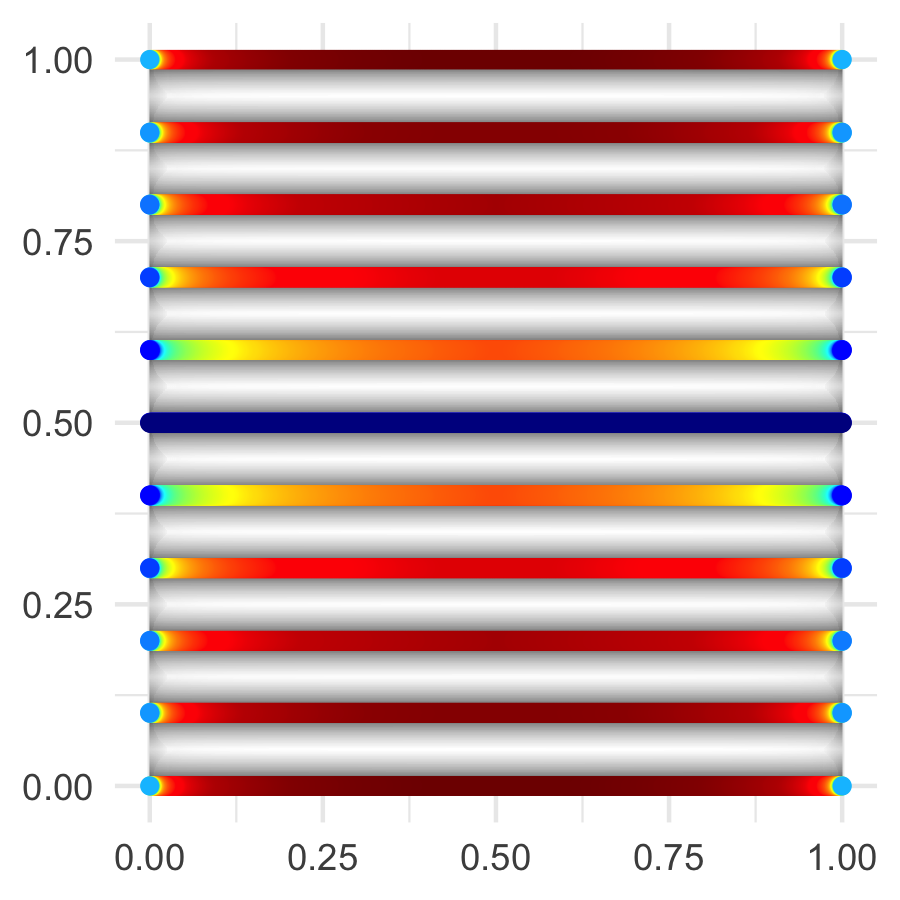}
    \includegraphics[width=0.32\textwidth, trim = 0mm 2mm 2mm 4mm, clip]{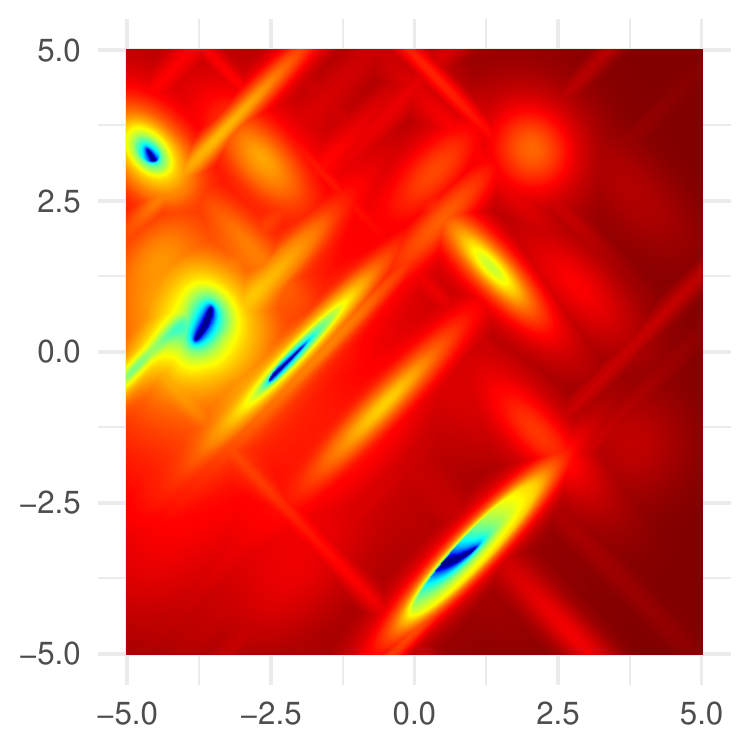}
    \includegraphics[width=0.32\textwidth, trim = 0mm 2mm 2mm 4mm, clip]{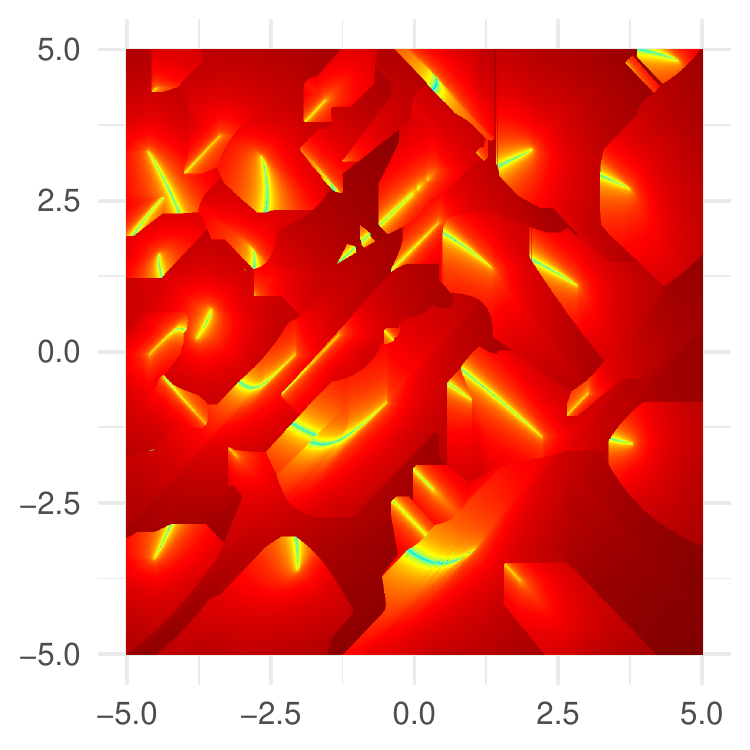}
    \includegraphics[width=0.32\textwidth, trim = 0mm 2mm 1mm 4mm, clip]{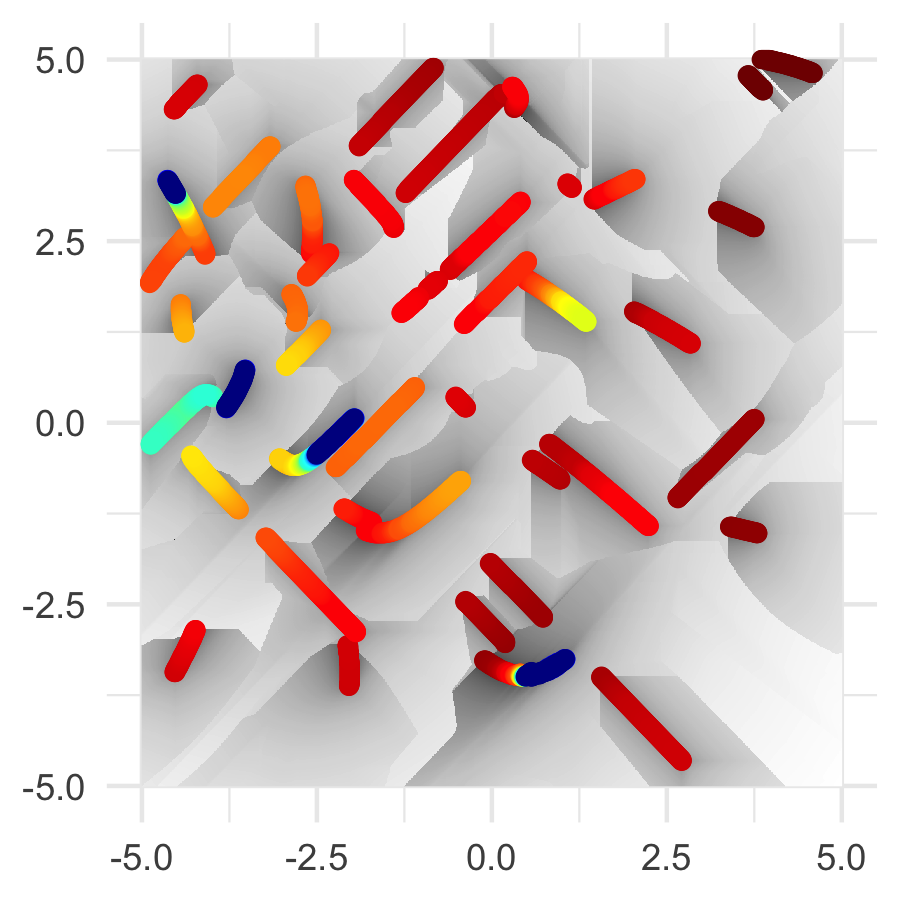}
    \caption{Comparison of the cost landscape, GFH and PLOT (left to right) on the two-dimensional bi-objective Aspar, DTLZ1 and bi-objective BBOB (FID = 10, IID = 1) functions (top to bottom). Due to the computational overhead involved in computing the domination counts, the cost landscape is calculated with only $500$ grid points per dimensions, while the GFH and PLOT use a resolution of $1,\!000$ points per dimension.}
    \label{fig:viz-comparison}
\end{figure} 

Fig.~\ref{fig:viz-comparison} provides a visual comparison of the current state-of-the-art visualization techniques -- cost landscapes (left column) and GFHs (middle column) -- and our proposed PLOT approach (right column) based on three exemplary MOPs: the simple Aspar function (top row) from Fig.~\ref{fig:aspar-gfh}, the well-established DTLZ~1~\cite{Deb2005} (middle row), and the 10th function from the rather recent bi-objective BBOB test suite~\cite{tusar2016}. Noticeably, for the latter two problems, GFH has problems in identifying some critical points correctly and mistakenly shows some points along the boundary as locally efficient. On the other hand, the cost landscape approach has problems with \emph{local} efficient sets and can at most identify the \emph{location} of some of them -- as long as their fronts are close to the global Pareto front(s). PLOT combines the global view of the cost landscapes with the local information of the GFH, and thus provides a much more informative depiction of the locally efficient solutions.

Our R-package \texttt{moPLOT}, which has been used to generate all visualizations in this paper, is available on GitHub: \url{https://github.com/kerschke/moPLOT}. Further resources and information can be found on our project's website on multimodal multi-objective optimization: \url{https://mo-opt.github.io}.

\section{Observations}
\label{sec:evaluation}

\begin{figure}[t!]
    \centering
    \includegraphics[width=0.32\textwidth, trim = 0mm 2mm 2mm 12mm, clip]{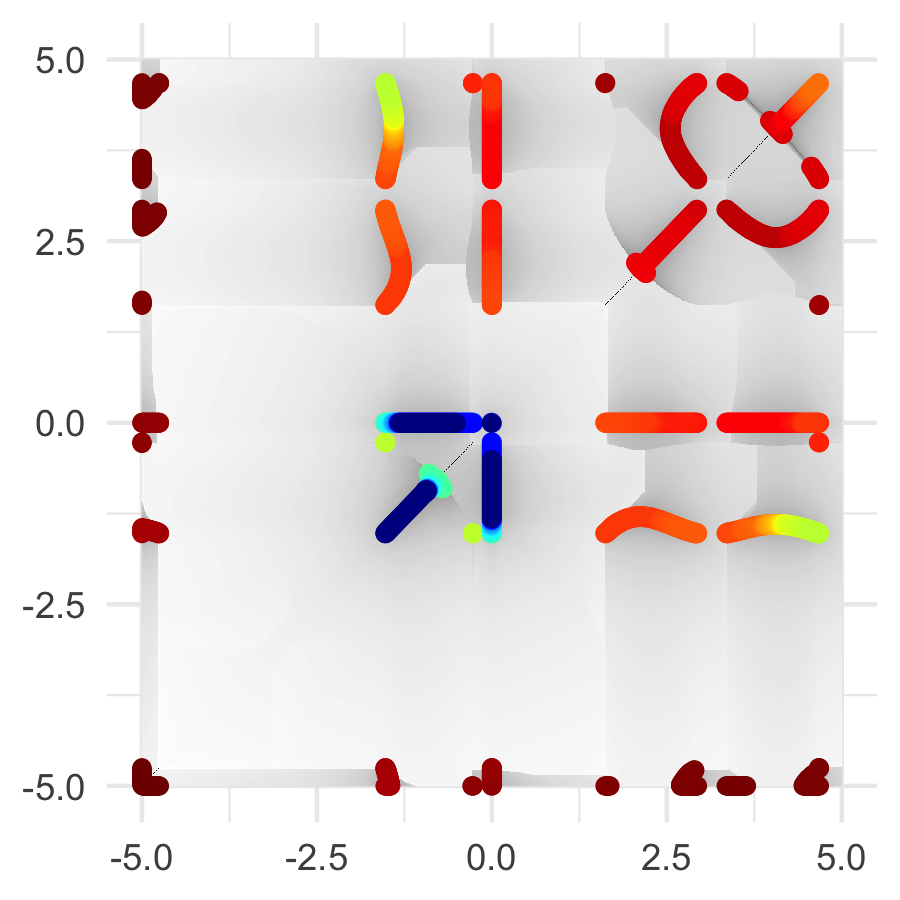}
    \includegraphics[width=0.32\textwidth, trim = 0mm 2mm 2mm 12mm, clip]{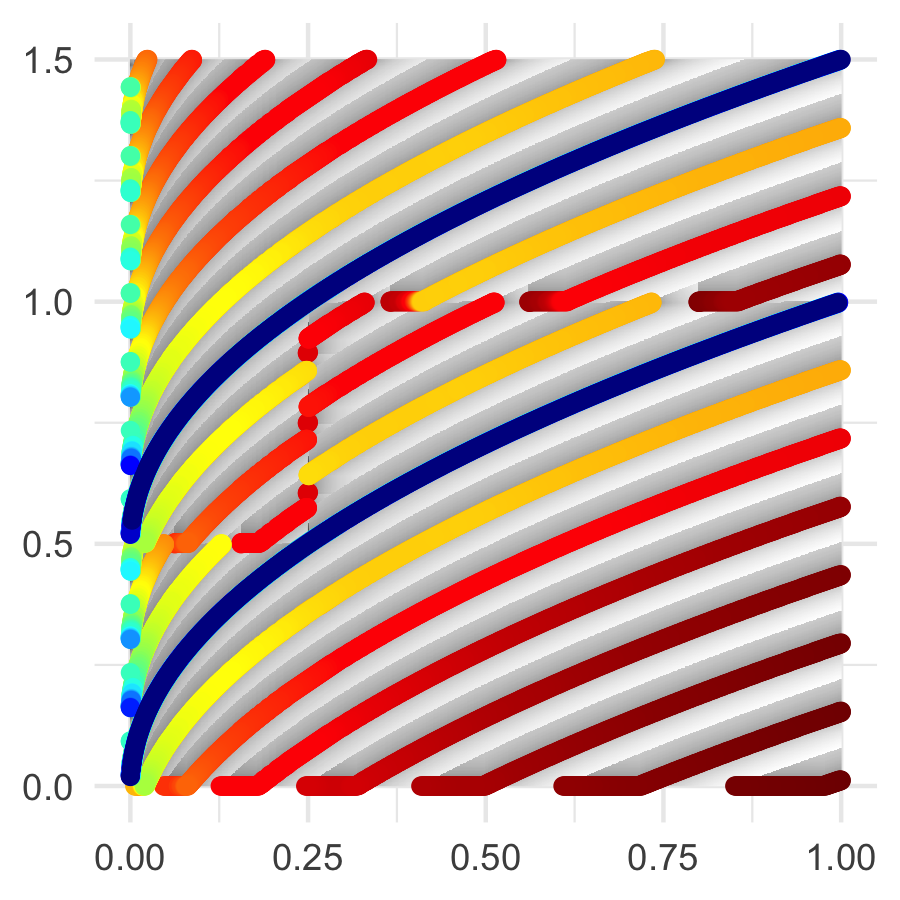}
    \includegraphics[width=0.32\textwidth, trim = 0mm 2mm 2mm 12mm, clip]{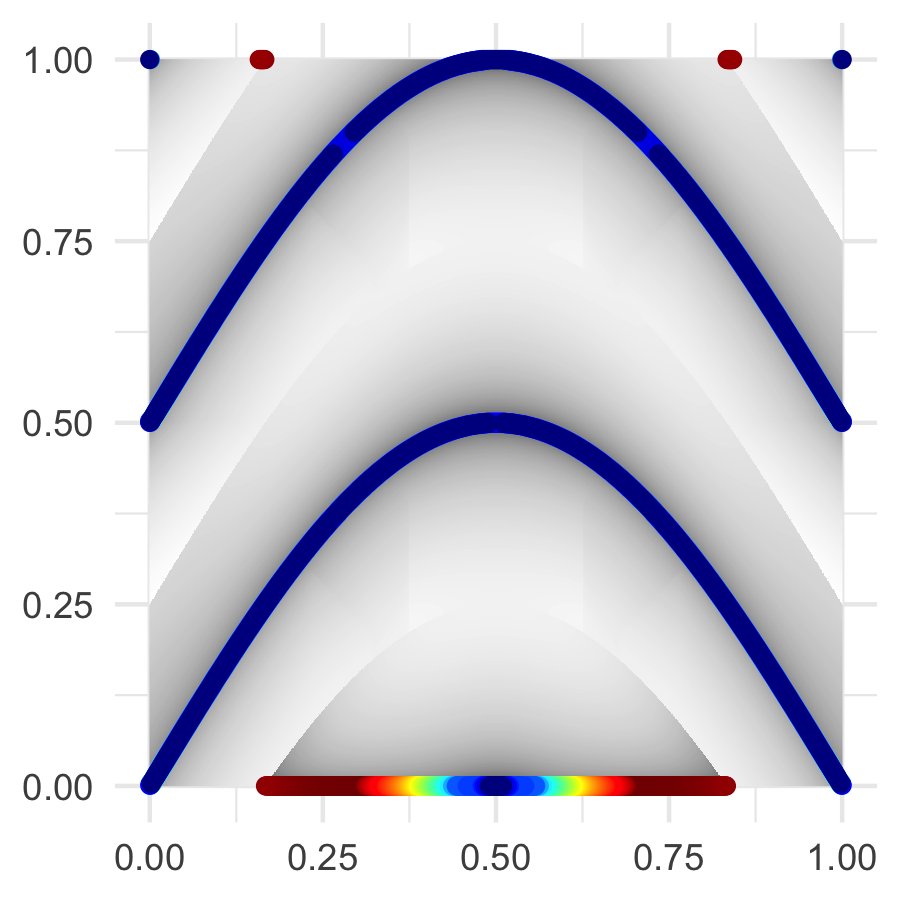}
    \includegraphics[width=0.32\textwidth, trim = 0mm 2mm 2mm 12mm, clip]{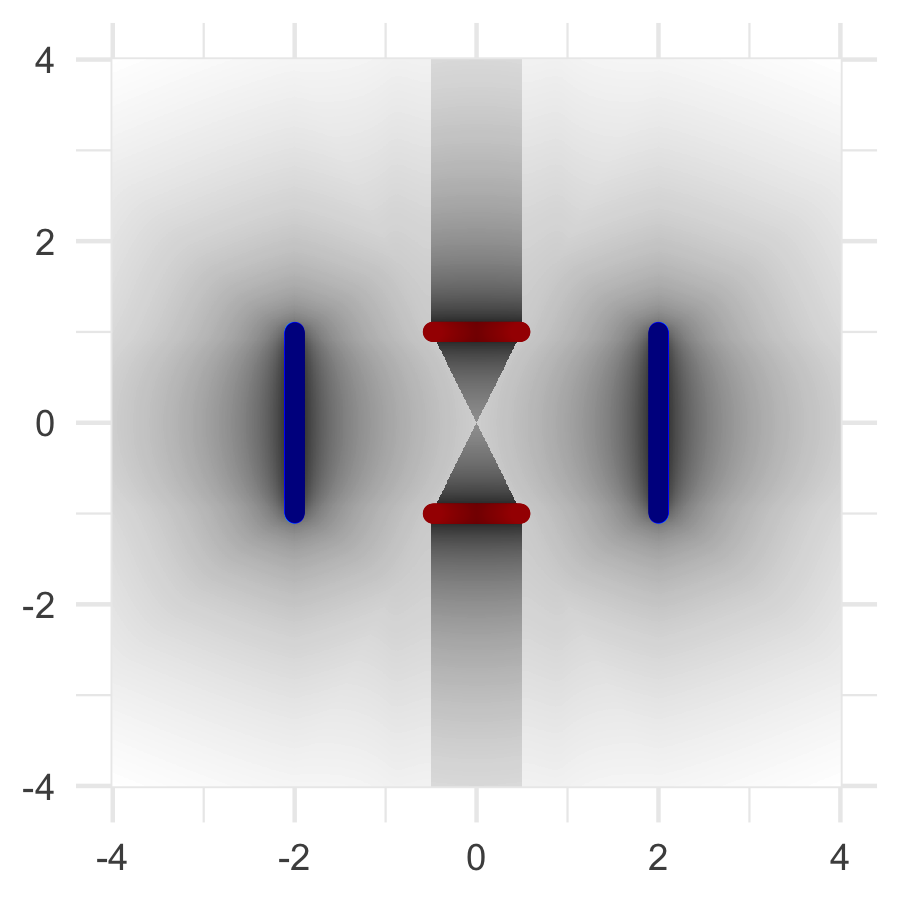}
    \includegraphics[width=0.32\textwidth, trim = 0mm 2mm 2mm 12mm, clip]{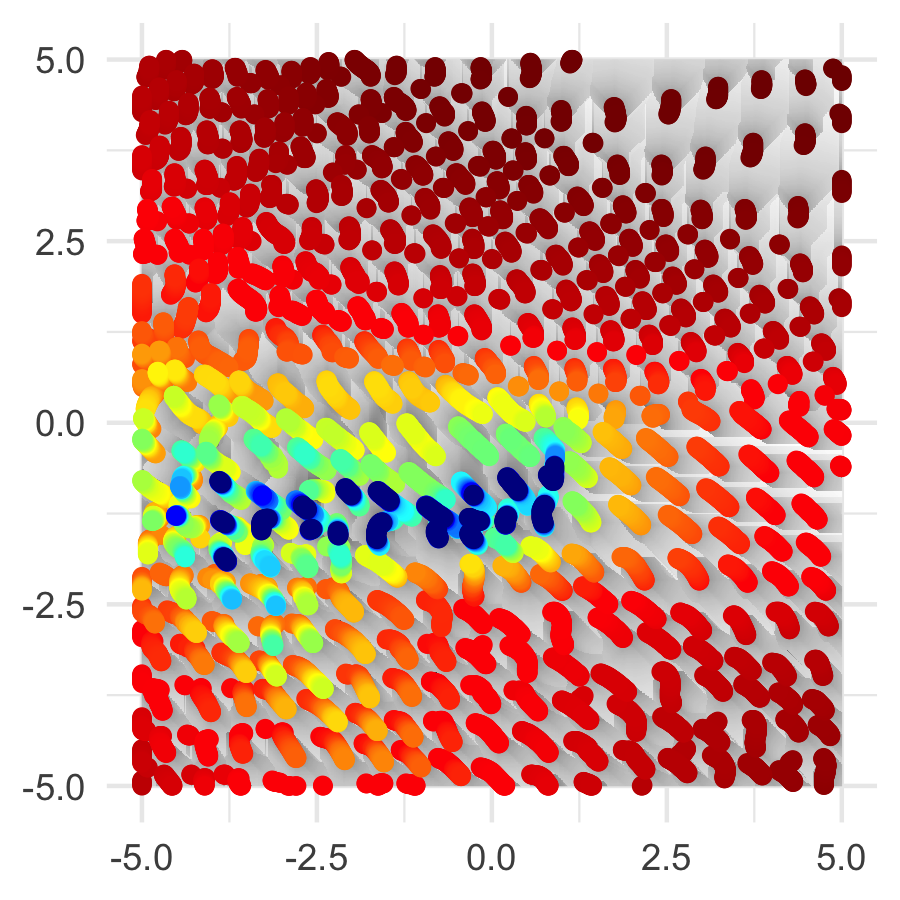}
    \includegraphics[width=0.32\textwidth, trim = 0mm 2mm 2mm 12mm, clip]{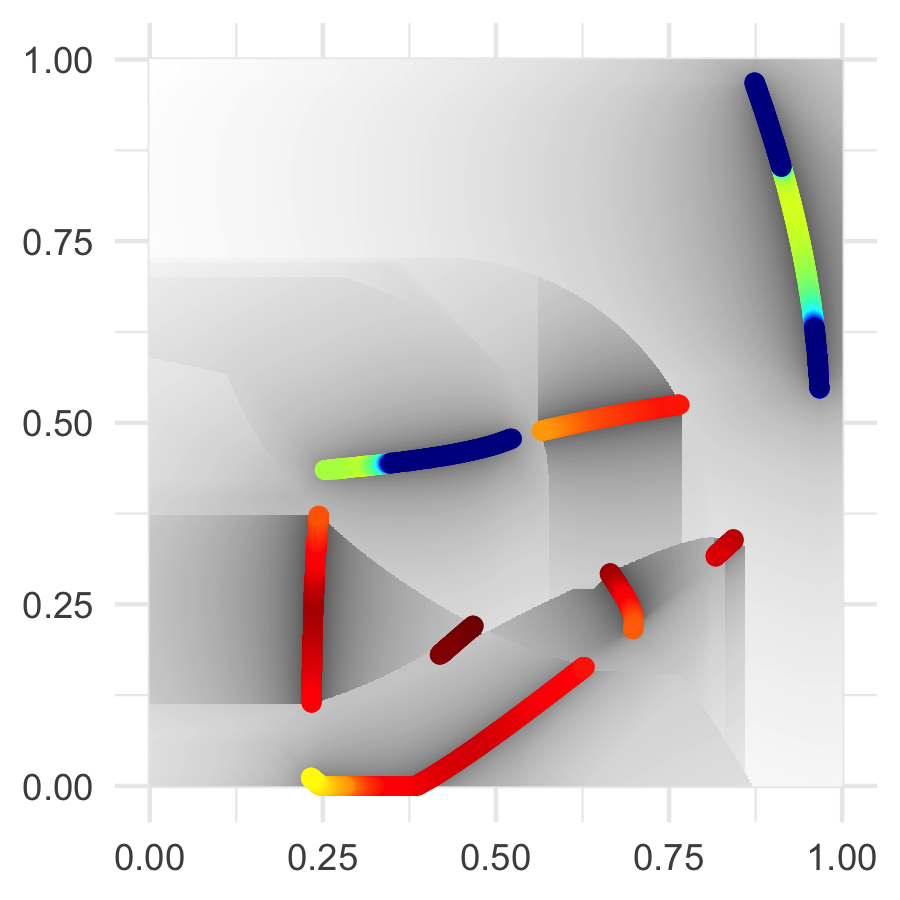}
    \caption{Exemplary PLOTs for various continuous MOPs (left to right, top to bottom): Kursawe~\cite{kursawe1990}, MMF3 and MMF14a from the CEC 2019 test suite \cite{yue2019novel}, a MinDist function with centers $(-2, -1), (2, 1)$ and $(-2, 1), (2, -1)$ \cite{maree2019}, a bi-objective BBOB (FID: 42, IID: 1) \cite{tusar2016}, and a bi-objective function generated using the MPM2 generator (with parameters $(3, 2, \text{random}, 4)$ and $(3, 2, \text{random}, 8)$) \cite{KerschkeWPGDTE16}. All plots were generated using an equidistant grid in the decision space with a resolution of $1,\!000 \times 1,\!000$ points.}
    \label{fig:viz-evaluation}
\end{figure} 

We provide PLOT visualizations for a selection of further benchmark functions in Fig.~\ref{fig:viz-evaluation}. Many MOPs that were designed with multimodality in mind reveal very simple structures in the decision space. Notably, the PLOTs show peculiarities in the definition of some functions that were designed with a focus on multiple global Pareto sets. These can lead to unintended locally and globally efficient solutions along the boundary (MMF14a) and glaring cuts in the landscape of the local efficient sets (MMF3). Otherwise, many MOPs have even simpler landscape structures only containing few local efficient sets in general (MinDist). Only few of the MOPs show interactions between the objectives, which lead to a disconnected global Pareto set (i.e., it is distributed over multiple local efficient sets). This can, e.g., be seen in the bi-objective BBOB and MPM2 functions.

Further note that the location of locally efficient points along the decision boundary implies that in general an unconstrained locally efficient set would be found outside of the feasible decision space. This can be observed in the depicted Kursawe, MMF and MPM2 functions.

The bi-objective BBOB function shows a very complex landscape with many locally efficient solutions. In fact, its sets cover the majority of the decision space and thereby reveal the limitations of PLOT. However, such extremely multimodal MOPs are challenging for any visualization method. Also, even for that very extreme problem, PLOT reliably visualized the MOP's global structure.

\section{Conclusions}
\label{sec:conclusion}

Visualizing an optimization problem's landscape is very useful when studying its properties, or the search behavior of the optimizers operating on it. In MO continuous optimization, however, there exist hardly any meaningful visualization methods, with the consequence that MOPs are primarily treated as black-boxes.

We present a novel approach for the numerical approximation of locally efficient points in the decision space of continuous MOPs. This new information was then integrated into PLOT -- our new method for the visualization of bi-objective two-dimensional MOPs. Our approach can visualize local and global efficient sets, as well as their basins of attraction. Thereby, it enables a visualization of MOPs that encompasses information comparable to visualizations available for single-objective functions. 
We successfully apply our approach to a wide variety of benchmarking functions and often reveal very simple landscape properties.
As with previous MO visualization techniques, we hope to inspire further progress in understanding the landscapes of existing benchmarking functions, the design of new benchmarks, as well as the development of novel algorithmic ideas.

It should be noted that the definitions for the MO gradient can be extended to an arbitrary number of dimensions and objectives \cite{desideri2012multiple}. 
Likewise, our approach for identifying critical points as well as our second-order criterion, which is based on the stability of the MO gradient field, can easily be adapted to cope well with higher-dimensional MOPs.
Thus, our proposed approach provides the fundamentals for an extension towards visualizing decision spaces of 3-dimensional MOPs. To the best of our knowledge, this is not yet available 
beyond the visualization of Pareto sets, or analytically known local efficient sets. 
An extension to 3D decision spaces would also enable more detailed investigations of the properties of 3-objective MOPs. This is currently not yet feasible, as their counterparts with 2D decision spaces in general contain degenerated critical points. %\ls{Noch ein Versuch \smiley{}}

Further, it can be noted that our approach supports studying selected regions of interest in the landscape. This can be effectively achieved by \emph{zooming} into the PLOT and supports the visualization of particularly complex MOPs. 
Another possible extension that aims at improving the visualization quality of our PLOT would be a dynamic resampling strategy around the identified critical points, increasing the accuracy of the approximation of the locally efficient points.

\subsection*{Acknowledgments}
The authors acknowledge support by the \href{https://www.ercis.org}{\textit{European Research Center for Information Systems (ERCIS)}}.

\bibliographystyle{unsrt}
\bibliography{arxiv}

\begin{thebibliography}{10}

\bibitem{Bey01}
H.-G. Beyer.
\newblock {\em {The Theory of Evolution Strategies}}.
\newblock Springer, 2001.

\bibitem{CC07}
Carlos~Artemio {Coello Coello}, David~A. {van Veldhuizen}, and Gary~B. Lamont.
\newblock {\em {Evolutionary Algorithms for Solving Multi-Objective Problems}}.
\newblock Springer, 2 edition, 2007.

\bibitem{Fonseca1995}
Carlos Manuel~Mira {da Fonseca}.
\newblock {\em {Multiobjective Genetic Algorithms with Application to Control
  Engineering Problems}}.
\newblock {PhD Thesis}, Department of Automatic Control and Systems
  Engineering, University of Sheffield, September 1995.

\bibitem{kerschke2017expedition}
Pascal Kerschke and Christian Grimme.
\newblock {An Expedition to Multimodal Multi-Objective Optimization
  Landscapes}.
\newblock In Heike Trautmann, G\"unter Rudolph, Klamroth Kathrin, Oliver
  Sch\"utze, Margaret Wiecek, Yaochu Jin, and Christian Grimme, editors, {\em
  {Proceedings of the 9th International Conference on Evolutionary
  Multi-Criterion Optimization (EMO)}}, pages 329~--~343. Springer, March 2017.

\bibitem{GrimmeKT2019Multimodality}
Christian Grimme, Pascal Kerschke, and Heike Trautmann.
\newblock {Multimodality in Multi-Objective Optimization --- More Boon than
  Bane?}
\newblock In {\em Evolutionary Multi-Criterion Optimization}, pages 126~--~138.
  Springer, 2019.

\bibitem{john2014extremum}
Fritz John.
\newblock {Extremum Problems with Inequalities as Subsidiary Conditions}.
\newblock In {\em {Traces and Emergence of Nonlinear Programming}}, pages
  197~--~215. Springer, 2014.

\bibitem{miettinen2012nonlinear}
Kaisa Miettinen.
\newblock {\em Nonlinear Multiobjective Optimization}, volume~12 of {\em
  {International Series in Operations Research \& Management Science}}.
\newblock Springer, 1998.

\bibitem{Custodio2018}
A.~L. Cust\'odio and J.~F.~A. Madeira.
\newblock {MultiGLODS: Global and Local Multiobjective Optimization Using
  Direct Search}.
\newblock {\em {Journal of Global Optimization}}, 72(2):323~--~345, 2018.

\bibitem{liefooghe:hal-01823666}
Arnaud Liefooghe, Manuel L{\'o}pez-Ib{\'a}{\~n}ez, Lu{\'i}s Paquete, and
  S{\'e}bastien Verel.
\newblock {Dominance, Epsilon, and Hypervolume Local Optimal Sets in
  Multi-Objective Optimization, and How to Tell the Difference}.
\newblock In {\em {Proceedings of the 20th Annual Conference on Genetic and
  Evolutionary Computation (GECCO)}}, volume~18, pages 324~--~331, Kyoto,
  Japan, 2018. {ACM}.

\bibitem{KerschkeWPGDTE16}
Pascal Kerschke, Hao Wang, Mike Preuss, Christian Grimme, Andr{\'e}~H. Deutz,
  Heike Trautmann, and Michael T.~M. Emmerich.
\newblock {Towards Analyzing Multimodality of Multiobjective Landscapes}.
\newblock In {\em {Proceedings of the 14th International Conference on Parallel
  Problem Solving from Nature ({PPSN XIV})}}, pages 962~--~972. Springer, 2016.

\bibitem{GrimmeKEPDT2019SlidingToThe}
Christian Grimme, Pascal Kerschke, Michael T.~M. Emmerich, Mike Preuss,
  Andr{\'e}~H. Deutz, and Heike Trautmann.
\newblock {Sliding to the Global Optimum: How to Benefit from Non-Global Optima
  in Multimodal Multi-Objective Optimization}.
\newblock In {\em {AIP Conference Proceedings}}, pages 020052--1--020052--4.
  AIP Publishing, 2019.

\bibitem{whitley1995building}
L.~Darrell Whitley, Keith~E. Mathias, Soraya~B. Rana, and John Dzubera.
\newblock {Building Better Test Functions}.
\newblock In {\em {Proceedings of the 6th International Conference on Genetic
  Algorithms (ICGA)}}, pages 239~--~247, 1995.

\bibitem{vanveldhuizen1999}
David~A. {van Veldhuizen}.
\newblock {\em {Multiobjective Evolutionary Algorithms: Classifications,
  Analyzes, and New Innovations}}.
\newblock PhD thesis, {Faculty of the Graduate School of Engineering of the Air
  Force Institute of Technology, Air University}, June 1999.

\bibitem{Zitzler2000}
Eckart Zitzler, Kalyanmoy Deb, and Lothar Thiele.
\newblock {Comparison of Multiobjective Evolutionary Algorithms: Empirical
  Results}.
\newblock {\em {Evolutionary Computation (ECJ)}}, 8(2):173~--~195, 2000.

\bibitem{Deb2005}
Kalyanmoy Deb, Lothar Thiele, Marco Laumanns, and Eckart Zitzler.
\newblock {Scalable Test Problems for Evolutionary Multiobjective
  Optimization}.
\newblock In {\em {Evolutionary Multiobjective Optimization}}, pages
  105~--~145. Springer, 2005.

\bibitem{tusar2016}
Tea Tu\v{s}ar, Dimo Brockhoff, Nikolaus Hansen, and Anne Auger.
\newblock {COCO: The Bi-Objective Black Box Optimization Benchmarking
  (bbob-biobj) Test Suite}.
\newblock {\em {arXiv preprint}}, abs/1604.00359, 2016.

\bibitem{yue2019novel}
Caitong Yue, Boyang Qu, Kunjie Yu, Jing Liang, and Xiaodong Li.
\newblock {A Novel Scalable Test Problem Suite for Multimodal Multiobjective
  Optimization}.
\newblock {\em {Swarm and Evolutionary Computation}}, 2019.

\bibitem{Tusar14phd}
Tea Tu{\v s}ar.
\newblock {\em {Visualizing Solution Sets in Multiobjective Optimization}}.
\newblock PhD thesis, Jo{\v z}ef Stefan International Postgrad.~School, 2014.

\bibitem{Tusar15tevc}
Tea Tu{\v s}ar and Bogdan Filipi{\v c}.
\newblock {Visualization of Pareto Front Approximations in Evolutionary
  Multiobjective Optimization: A Critical Review and the Prosection Method}.
\newblock {\em {IEEE Transactions in Evolutionary Computation (TEVC)}},
  19(2):225~--~245, 2015.

\bibitem{kerschke2018search}
Pascal Kerschke, Hao Wang, Mike Preuss, Christian Grimme, Andr{\'e}~H. Deutz,
  Heike Trautmann, and Michael T.~M. Emmerich.
\newblock {Search Dynamics on Multimodal Multi-Objective Problems}.
\newblock {\em {Evolutionary Computation (ECJ)}}, 27:577~--~609, 2019.

\bibitem{Deb99}
Kalyanmoy Deb.
\newblock {Multi-Objective Genetic Algorithms: Problem Difficulties and
  Construction of Test Problems}.
\newblock {\em {Evolutionary Computation (ECJ)}}, 7(3):205~--~230, 1999.

\bibitem{blanchard2012differential}
P.~Blanchard, R.L. Devaney, and G.R. Hall.
\newblock {\em Differential Equations}.
\newblock Cengage Learning, 2012.

\bibitem{kursawe1990}
Frank Kursawe.
\newblock {A Variant of Evolution Strategies for Vector Optimization}.
\newblock In {\em {Proceedings of the 1st International Conference on Parallel
  Problem Solving from Nature (PPSN I)}}, pages 193~--~197. Springer, 1990.

\bibitem{maree2019}
Stef~C. Maree, Tanja Alderliesten, and Peter A.~N. Bosman.
\newblock {Real-Valued Evolutionary Multi-Modal Multi-Objective Optimization by
  Hill-Valley Clustering}.
\newblock In {\em {Proceedings of the 21st Annual Conference on Genetic and
  Evolutionary Computation (GECCO)}}, pages 568~--~576. ACM, 2019.

\bibitem{desideri2012multiple}
Jean-Antoine D{\'e}sid{\'e}ri.
\newblock {Multiple-Gradient Descent Algorithm (MGDA) for Multiobjective
  Optimization}.
\newblock {\em Comptes Rendus Mathematique}, 350(5-6):313~--~318, 2012.

\end{thebibliography}

\end{document}